\pdfoutput=1
\documentclass{article}

\usepackage{PRIMEarxiv}

\usepackage[utf8]{inputenc} % allow utf-8 input
\usepackage[T1]{fontenc}    % use 8-bit T1 fonts
\usepackage{url}            % simple URL typesetting
\usepackage{booktabs}       % professional-quality tables
\usepackage{amsfonts}       % blackboard math symbols
\usepackage{nicefrac}       % compact symbols for 1/2, etc.
\usepackage{microtype}      % microtypography
\usepackage{fancyhdr}       % header
\usepackage{graphicx}       % graphics
\usepackage{xcolor}
\usepackage[colorlinks=true, allcolors=blue]{hyperref} % hyperlinks

\usepackage{float}
\usepackage{amsmath,amssymb}
\usepackage{array}
\usepackage{multirow}
\usepackage{siunitx}
\usepackage{tabularx}
\usepackage{subcaption}
\usepackage[section]{placeins}

\let\oldtabular\tabular
\renewcommand{\tabular}{\footnotesize\oldtabular}

\let\oldthebibliography\thebibliography
\renewcommand{\thebibliography}[1]{%
  \oldthebibliography{#1}%
  \setlength{\itemsep}{0pt}%
  \setlength{\parsep}{0pt}%
  \small}

\newcommand{\Photdr}{\ensuremath{\Phi}\text{-OTDR}}
\newcommand{\pmv}[2]{\mbox{$#1 \pm #2\%$}}
\newcommand{\pmvb}[2]{\mbox{$\mathbf{#1 \pm #2\%}$}}
\providecommand{\doi}[1]{\url{https://doi.org/#1}}
\newcommand{\orcid}[1]{\href{https://orcid.org/#1}{\includegraphics[width=8pt]{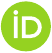}}}
\title{Vision-Language Model-Based Phase-OTDR Event Classification
}

\author{
  Ahmet Bestami Kose$^{1}$\,\orcid{0000-0003-3817-0999} \qquad
  Mehmet Sergen Catal$^{2}$\,\orcid{0000-0002-8501-6495} \qquad
  Muhammet Cagri Yeke$^{3}$\,\orcid{0009-0002-6896-1724} \\
  \bfseries Kivilcim Yuksel$^{2}$\,\orcid{0000-0003-1512-3022} \qquad
  Abdurrahman Gumus$^{1,*}$\,\orcid{0000-0003-2993-5769} \\[8pt]
  \normalfont\small $^{1}$Department of Computer Engineering, Isparta University of Applied Sciences, 32040 Isparta, Turkey \\
  \normalfont\small $^{2}$Department of Electrical and Electronics Engineering, Izmir Institute of Technology, 35430 Urla, Izmir, Turkey \\
  \normalfont\small $^{3}$Department of Biotechnology and Bioengineering, Izmir Institute of Technology, 35430 Urla, Izmir, Turkey \\[4pt]
  \normalfont\small $^{*}$Corresponding author: \texttt{abdurrahmangumus@isparta.edu.tr}
}

\begin{document}
\maketitle

\begin{abstract}
Phase-sensitive optical time-domain reflectometry ($\Phi$-OTDR) enables distributed
vibration sensing, but reliable event classification has traditionally relied on fully
supervised models that require extensive annotated data. This study examines the
adaptability of a Vision-Language Model (VLM) for $\Phi$-OTDR event recognition through
few-shot learning. Raw samples are converted into Short-Time Fourier Transform (STFT)
dB-magnitude spectrogram mosaics, color-mapped and resized to RGB images, and
processed by a Contrastive Language--Image Pre-training (CLIP) image encoder;
few-shot adaptation optimizes only the Tip-Adapter-F cache keys, leaving both encoders unchanged. Under a leakage-controlled protocol (train-side support/validation pool of 512
images per class, fixed held-out test of 120 per class, no test-set access during
selection, 20 distinct support-sampling trials), the selected configuration
(the ViT-L-14 Vision Transformer encoder with OpenAI weights, 128 shots per class) reaches
\pmv{96.81}{0.42} accuracy and a macro-averaged F1 score of \pmv{96.79}{0.42} on the held-out test set; a label-efficient 16-shot setting, that is, 96 labeled
support images at adaptation time across the six classes, already attains
\pmv{90.17}{0.81}. Tip-Adapter-F clearly outperforms both zero-shot CLIP and the
standard training-free Tip-Adapter, and prompt and backbone analyses show that performance
is stable across Vision Transformer encoders and insensitive to prompt wording at the selected operating point. Overall, CLIP visual features carry a transferable structure that a lightweight
support cache turns into a label-efficient $\Phi$-OTDR event classifier. The code and the
corresponding image dataset are made publicly available on GitHub to support further research:
\url{https://github.com/miralab-ai/vlm-phase-otdr}.
\end{abstract}

% keywords can be removed
\keywords{$\Phi$-OTDR \and Event classification \and Vision-language models \and Few-shot learning \and CLIP}

\section{Introduction}

Distributed fiber-optic sensing based on $\Phi$-OTDR has emerged as a crucial technology for long-range perimeter monitoring, infrastructure surveillance, pipeline security, and vibration event recognition. A $\Phi$-OTDR interrogator captures variations in Rayleigh backscattering along the fiber, generating high-dimensional spatio-temporal signals whose patterns are strictly dictated by the physical disturbances acting upon the sensing cable. Although deep learning has significantly improved event recognition accuracy, most existing models rely on conventional supervised learning, which requires extensive annotated datasets. This reliance severely restricts their adaptability in real-world deployments, where variations in soil conditions, deployment geometries, cable coupling, or novel event types often emerge with only a minimal number of labeled examples available.

From a measurement perspective, event classification is the final stage of the \Photdr{} measurement chain: the interrogator records the backscattered intensity, and the classifier converts it into the quantity on which an operator acts, namely which disturbance occurred at a given position along the fiber. The figure of merit of this stage is its classification accuracy, and its practical cost is set largely by the number of labeled recordings that must be collected at each new installation site or for each new event type. A measurement-oriented evaluation should therefore report accuracy as a function of this labeling budget, state the uncertainty of the reported accuracy, and prevent any information from the test recordings from entering model selection. The present study is designed around these three requirements.

To address this limitation, this study formulates $\Phi$-OTDR event recognition as a signal-to-image few-shot adaptation problem. Rather than training a task-specific vision model from scratch, we transform each one-dimensional, multi-position $\Phi$-OTDR sample into a spectrogram mosaic, which is then processed by a pre-trained CLIP image encoder. The few-shot adaptation is subsequently executed using Tip-Adapter and Tip-Adapter-F, integrating text-prompt logits with evidence from a support-set cache. Ultimately, the core research question is not whether zero-shot CLIP can directly classify $\Phi$-OTDR events, but rather if its visual representation can provide a robust, transferable feature space for signal-derived imagery when guided by a small labeled cache. Throughout, the task is a six-class
\emph{event classification} problem (one of which is a no-event
\emph{Background} category), and we use ``event classification'' consistently.
\subsection{Related Work}

Early recognition pipelines combined hand-crafted time-- and frequency-domain descriptors with conventional classifiers. Cao \textit{et al.} introduced an open $\Phi$-OTDR benchmark with six event categories (Background, Digging, Knocking, Watering, Shaking, Walking) and two reference baselines: an engineered-feature Support Vector Machine (SVM) ($\approx\!82.6\%$) and a 2-D Convolutional Neural Network (CNN) ($\approx\!94.0\%$)~\cite{cao2023}. It has since become a widely adopted testbed. Subsequent work improved supervised accuracy with deeper architectures: a Dendrite-Network classifier ($\approx\!98.6\%$)~\cite{chen2023dendrite}, a semi-supervised Attention-based Convolutional Neural Network and Bidirectional Long Short-Term Memory~\cite{li2024multievent}, and transformer models including Inter-Sequence-Attention Transformer (ISAT)~\cite{duan2025isat} and the long-sequence Distributed fiber-optic acoustic sensing (DAS)-Former ($\approx\!99.6\%$)~\cite{li2025dasformer}. Further directions include transfer learning~\cite{gan2024transfer}, intensity--phase stacked deep transfer learning with recurrent networks~\cite{kayan2023}, event augmentation and Generative Adversarial Network (GAN) balancing~\cite{hu2024augment}, Mel-Frequency Cepstral Coefficients/differential-phase stacking~\cite{kamanga2024}, feature selection with gradient-boosting classifiers~\cite{sasi2024}, knowledge distillation~\cite{luo2025kd}, contrastive-prototype open-set recognition~\cite{wang2025irms}, and global-local masked-autoencoder pre-training~\cite{cheng2025glmae}. These methods perform strongly but assume abundant labels and are typically trained end-to-end on 8:2 splits. Label scarcity has begun to be addressed for $\Phi$-OTDR through augmentation-based few-shot learning~\cite{shi2022dualaug}, zero-shot event recognition~\cite{hu2024zeroshot}, attention-based prototypical few-shot networks~\cite{yu2025proto}, and rapid tabular few-shot classification on the same open benchmark~\cite{lin2025tabpfn}; however, adapting a fixed vision-language encoder through a support-set cache remains unexplored in this setting.

The most recent label-efficient work relies mainly on domain-specific self-supervised
  pre-training. Wang et al.\ combine
self-supervised spatiotemporal pre-training with a bidirectional Mamba2 backbone,
few-shot adaptation, and pseudo-label refinement, reporting $94.90\%$, $98.10\%$,
$98.63\%$ and $99.37\%$ accuracy at 1, 4, 8 and 16 shots per
class~\cite{wang2026ssbimamba}. Zeng et al.\ learn \Photdr{} signal representations by
self-supervised distillation and report $96.60\%$ accuracy from 15 labeled samples per
class~\cite{zeng2026ssd}. Working from the data side instead, Zhang and Xiao pair hybrid
augmentation (noise injection and time stretching together with a conditional GAN) with a
pre-trained VGGish encoder, a BiLSTM and an attention module, and report $96.2\%$ accuracy
and $96.1\%$ F1 from 70 training samples per class on the same open
benchmark~\cite{zhang2026vba}.

These figures are not directly comparable with the few-shot results reported here,
because the supervision budgets differ in kind rather than in degree. The self-supervised
approaches of Wang et al.~\cite{wang2026ssbimamba} and Zeng et al.~\cite{zeng2026ssd}
first pre-train a task-specific backbone on the unlabeled remainder of the benchmark's own training partition, so the labeled shots they quote are spent on top of a representation that has already seen the sensing domain; the augmentation-based model of Zhang and Xiao~\cite{zhang2026vba} likewise trains its temporal encoder end-to-end on the target
data. The present study starts from neither: no \Photdr{} data, labeled or unlabeled, is
used to pre-train or fine-tune the encoders, which remain exactly as released, and the only
learned parameters are the $6K$ support-cache keys. Table~\ref{tab:regimes} places the two
regimes side by side. Domain-specific self-supervised pre-training buys higher few-shot
accuracy at the cost of an unlabeled in-domain corpus and a full pre-training stage,
whereas the route taken here buys immediate deployability from an off-the-shelf
large-scale vision-language backbone plus a very light task adaptation. The question
addressed in what follows is therefore not which number is larger, but how much of the
\Photdr{} recognition problem a fixed, general-purpose vision-language representation
already solves.

\begin{table}[htbp]
\centering
\caption{Supervision budgets of recent label-efficient \Photdr{} event-recognition
methods on the open benchmark of Cao et al.~\cite{cao2023}. The accuracies are not
directly comparable: the rows differ in whether a \Photdr{}-specific representation is
learned before the labeled shots are counted. \emph{Unlab.\ data} states whether an
unlabeled in-domain \Photdr{} corpus is required for that prior stage.}
\label{tab:regimes}

\scriptsize
\setlength{\tabcolsep}{2.5pt}
\renewcommand{\arraystretch}{1.0}

\begin{tabularx}{\linewidth}{
    >{\raggedright\arraybackslash}p{0.175\linewidth}
    >{\raggedright\arraybackslash}X
    >{\centering\arraybackslash}p{0.070\linewidth}
    >{\centering\arraybackslash}p{0.075\linewidth}
    >{\raggedright\arraybackslash}p{0.180\linewidth}
    >{\centering\arraybackslash}p{0.065\linewidth}}
\toprule
Method
& Representation pre-training
& Unlab.\ data
& Shots / class
& Trained components
& Acc. \\
\midrule

2-D CNN~\cite{cao2023}
& none (random init.)
& no
& full split
& whole network
& $94.0\%$ \\

VBA-HA~\cite{zhang2026vba}
& VGGish (audio)
& no
& 70
& VGGish head, BiLSTM, attention
& $96.2\%$ \\

\mbox{SS-BiMamba}~\cite{wang2026ssbimamba}
& self-supervised on \Photdr{}
& yes
& 16
& encoder + MLP head; pseudo-labels
& $99.4\%$ \\

Self-distill.~\cite{zeng2026ssd}
& self-supervised on \Photdr{}
& yes
& 15
& encoder + head
& $96.6\%$ \\

\midrule
This study
& CLIP (web image--text), frozen
& no
& 16
& support-cache keys only
& $90.2\%$ \\

This study
& CLIP (web image--text), frozen
& no
& 128
& support-cache keys only
& $96.8\%$ \\

\bottomrule
\end{tabularx}
\end{table}

A common approach for adapting modern vision architectures to one-dimensional sensor signals involves transforming time-series measurements into two-dimensional images. Time--frequency transforms, such as the STFT and the continuous wavelet transform, capture transient and non-stationary spectral structures and have been directly applied to $\Phi$-OTDR sensing~\cite{qin2012cwt}. Another group of methods encodes temporal dependencies as spatial textures; for example, Gramian Angular Fields and Markov Transition Fields~\cite{wang2015gaf} (with a $\Phi$-OTDR application of the latter in~\cite{zhao2022mtf}), and Recurrence Plots~\cite{eckmann1987rp} have been utilized in distributed fiber sensing. Applications include Gramian-Angular-Field-based pipeline threat identification~\cite{li2023gaf}, image-coding-driven vibration detection~\cite{hu2024imagecoding}, and multi-transform image fusion (Gramian Angular Difference/Summation Fields and Recurrence Plots) with fine-tuned CNNs for $\Phi$-OTDR event detection~\cite{yeke2026}, with broader surveys confirming the general applicability of the signal-to-image paradigm across industrial diagnostics~\cite{garcia2022}. Across these signal-to-image approaches, the derived images serve as inputs to CNNs trained or fine-tuned under full supervision; they are not used as inputs to a pre-aligned vision-language feature space. Extending this line of research, the current work converts multi-position $\Phi$-OTDR samples into STFT dB-magnitude spectrogram mosaics that retain per-channel time--frequency signatures while arranging the 12 sensing channels in a compact sequential mosaic, and repurposes them as queries to a pre-trained CLIP encoder rather than as training data for a bespoke network.

CLIP aligns image and text embeddings from large-scale image--caption pairs, enabling zero-shot classification through text prompting~\cite{radford2021clip}. However, zero-shot CLIP often performs inadequately in specialized scientific or sensing domains, prompting the development of numerous parameter-efficient adaptation methods. Prompt-learning approaches optimize continuous context vectors; for example, CoOp learns class-shared prompts~\cite{zhou2022coop}, while CoCoOp conditions prompts on the input image to enhance generalization~\cite{zhou2022cocoop}. MaPLe instead prompts both branches jointly, conditioning vision prompts on language prompts to promote cross-modal synergy~\cite{khattak2023maple}. Adapter-style methods operate on precomputed embeddings: CLIP-Adapter introduces a lightweight residual feature adapter~\cite{gao2024clipadapter}, Tip-Adapter constructs a training-free key--value cache from the support set, and Tip-Adapter-F makes cache keys learnable while keeping the encoders fixed for improved label and computational efficiency~\cite{zhang2022tipadapter}. Recent advancements such as APE further separate domain-specific priors from the cache to improve few-shot discrimination~\cite{zhu2023ape}. In a related biomedical sensing setting, a CLIP-based few-shot Tip-Adapter approach has been applied to EEG connectivity-derived images for attention-deficit/hyperactivity disorder recognition~\cite{catal2025}, illustrating the promise of vision-language features on signal-derived imagery. Overall, the literature suggests that adapting CLIP features using a small support cache provides an effective approach to label-efficient recognition, although this strategy has been rarely evaluated on signal-derived imagery, particularly for distributed fiber-optic ($\Phi$-OTDR) sensing.

CLIP has only recently been applied to $\Phi$-OTDR; on the same open benchmark~\cite{cao2023}, Lin, Wu, and Boulet introduced a contrastive language-image pre-training methodology for $\Phi$-OTDR event classification~\cite{lin2025clipotdr}. Their approach adopts the CLIP training objective rather than the pre-trained CLIP model itself: the text branch is the frozen CLIP ViT-B-16 text encoder, whereas the visual branch is not the CLIP image encoder but the two-convolution-module CNN of the supervised baseline in~\cite{cao2023}, re-headed with a linear projection into the 512-dimensional text embedding space and trained from random initialization under a symmetric InfoNCE loss, that is, a contrastive objective that maximizes the similarity of matched image--text pairs while suppressing that of all mismatched pairs within the batch. Their model therefore contrastively trains this task-specific image encoder against a fixed text encoder to exploit the hierarchical main- and sub-label structure of the dataset, using the full training split and prompts of the form ``main: \{main-label\}, subclass: \{sub-label\}''; six-class predictions are subsequently obtained without retraining by truncating the prompt set to the main label alone. Two consequences follow for comparison purposes. First, no large-scale visual pre-training is transferred in~\cite{lin2025clipotdr}: the visual representation is learned entirely from $\Phi$-OTDR data, so the reported gains isolate the contribution of the contrastive text-alignment objective rather than that of a pre-trained vision-language representation. Second, their six-label result is obtained under full-data training rather than the few-shot regime adopted here, with supervision at the finer 25-sub-label granularity. By contrast, the present study keeps a large-scale pre-trained CLIP encoder entirely fixed in both modalities and adapts only a lightweight support cache using Tip-Adapter-F under a few-shot regime. This framework is further distinguished by a staged, leakage-controlled validation protocol, prompt and backbone ablations, few-shot scaling analysis, and class-pair error diagnostics. The primary contribution is not simply the application of CLIP to $\Phi$-OTDR data, but the formulation of a leakage-controlled few-shot VLM evaluation framework specifically designed for signal-derived $\Phi$-OTDR visual representations. 
 
\subsection{Literature Gaps}

Although supervised $\Phi$-OTDR event recognition has advanced considerably, several gaps motivate the present study. First, and most centrally, vision-language models, which jointly embed images and natural-language text in a shared representation space, remain almost entirely unexplored for $\Phi$-OTDR. Nearly all existing pipelines, from the SVM/CNN baselines of the open benchmark~\cite{cao2023} to recent transformer~\cite{li2025dasformer} and masked-autoencoder~\cite{cheng2025glmae} models and the few-shot, zero-shot and self-supervised label-efficient variants noted above, are unimodal: they map a signal-derived representation directly to a class index with no language component. This holds for the strongest current label-efficient results as well: both Wang et al.~\cite{wang2026ssbimamba} and Zeng et al.~\cite{zeng2026ssd} reach their few-shot accuracies by pre-training a unimodal \Photdr{} encoder on in-domain data, not by transferring a representation learned outside the sensing domain. To date, only a single CLIP-based $\Phi$-OTDR study exists~\cite{lin2025clipotdr}, and it trains a task-specific image encoder from scratch under a CLIP-style objective, using the full labeled training set, rather than transferring pre-aligned vision-language features, leaving the fixed-encoder, few-shot VLM regime for $\Phi$-OTDR essentially open.

Second, this unimodal framing overlooks what a VLM adds in principle over a plain
classifier: a language modality. Instead of emitting only a numeric label, a VLM matches
each measurement against textual descriptions of the events, so class semantics are
expressed as natural-language prompts and recognition becomes text-conditioned. In
principle this opens up capabilities that a conventional softmax classifier cannot offer:
semantically meaningful and editable class definitions, prompt design as a modeling axis,
and a natural route to extending or redefining the event set with few or even no
additional labeled examples. None of these has been systematically studied for
distributed fiber-optic sensing.

Third, signal-derived visual representations have been paired almost exclusively with
fully supervised CNNs and not evaluated with pre-trained vision-language features or
parameter-efficient few-shot mechanisms such as Tip-Adapter/Tip-Adapter-F~\cite{zhang2022tipadapter}
and its adaptive-prior refinement~\cite{zhu2023ape},
particularly under the label-scarce, rapid-adaptation conditions that field deployment
demands. The interplay of prompt design, CLIP backbone, and residual error structure is
likewise seldom analyzed under a strict validation/test separation, making it hard to
distinguish genuine adaptation from methodological leakage.

\subsection{Motivation and Study Outline}

This study aims to minimize annotation overhead while retaining the operational
reliability that fiber-optic monitoring demands, and to examine what the language
modality of a VLM contributes to $\Phi$-OTDR recognition. Rather than training a
task-specific network end-to-end to emit a bare class index, we cast each event as a
natural-language concept and ask whether a pre-trained CLIP encoder supplies a
transferable, text-alignable feature space for $\Phi$-OTDR spectrogram mosaics, and
whether a lightweight few-shot cache converts it into a robust six-class classifier
under a fraction of the labeled data and a leakage-controlled protocol. This provides
a label-efficient, text-conditioned alternative to the fully supervised, image-based CNN
pipelines previously used for $\Phi$-OTDR and, unlike the single prior CLIP-based study,
which trains a task-specific image encoder from scratch under a CLIP-style
objective~\cite{lin2025clipotdr}, keeps the pre-trained vision-language backbone
entirely fixed. Sections~\ref{sec:dataset}--\ref{sec:fewshot} describe the dataset, the
mosaic-generation pipeline, and CLIP/Tip-Adapter-F; Section~\ref{sec:experiments} reports
the staged validation and held-out evaluation (method comparison, few-shot scaling,
prompt and backbone analyses, class-wise diagnostics, and measurement uncertainty);
Section~\ref{sec:conclusion} concludes.

\subsection{Novelties and Contributions}

The study addresses the three gaps above by treating $\Phi$-OTDR event recognition as a
few-shot image--text matching task rather than a supervised signal-to-label task. The
pre-trained CLIP encoders are kept fixed in both modalities, adaptation is restricted to a
lightweight support cache, and every design choice is fixed on train-side data before the
held-out test set is touched. On this basis the main contributions are the following.
\begin{itemize}
  \item A vision-language formulation for \Photdr{}, in which each event is expressed as a
  natural-language prompt and scored against a pre-trained CLIP encoder over spectrogram
  mosaics. This brings the language modality, and prompt design as an explicit modeling
  axis, to distributed fiber-optic sensing, in contrast to the unimodal signal-to-label
  CNNs of prior work.
  \item A leakage-controlled few-shot framework that combines spectrogram mosaics with a
  fixed CLIP encoder and Tip-Adapter-F under a staged protocol: backbone screening,
  hyperparameter optimization, and held-out evaluation over 20 support-sampling trials. To the best of our knowledge, this is the first few-shot vision-language study of $\Phi$-OTDR event recognition, and the first to adapt a frozen, general-purpose vision-language backbone to this task without any $\Phi$-OTDR pre-training or fine-tuning of its encoders: the single prior CLIP-based study trains its image encoder from scratch on the full labeled set~\cite{lin2025clipotdr}, while the strongest recent label-efficient models first pre-train a backbone on an unlabeled in-domain corpus~\cite{wang2026ssbimamba} or distil one from it ~\cite{zeng2026ssd}.
  \item A benchmark evaluation on the open $\Phi$-OTDR dataset~\cite{cao2023}, comparing the
  adapted model against zero-shot CLIP and the training-free Tip-Adapter, together with a
  class-pair error analysis that identifies the physically interpretable
  Shaking--Watering confusion as the dominant residual error.
  \item Scaling and robustness analyses over shot count, prompt family, and CLIP backbone,
  showing that the approach is label-efficient (roughly $90\%$ accuracy at 16 shots) and
  stable across Vision Transformer encoders and prompt designs at high shot counts.
  \item A measurement-oriented evaluation that reports classification accuracy as a function of the labeling budget, from 1 to 128 labeled recordings per class, together with an uncertainty budget that separates the variability due to the choice of support recordings from the uncertainty due to the finite test set (section~\ref{sec:uncertainty}).
\end{itemize}

\section{Materials and Methods}
\subsection{$\Phi$-OTDR Dataset}
\label{sec:dataset}

The experiments use the publicly available \Photdr{} event dataset introduced by
Cao et al.~\cite{cao2023}, acquired with an intensity-based (direct-detection) \Photdr{}
system that records vibration-induced variations of the Rayleigh-backscattered intensity
along an optical fiber. The sensing fiber consists of a bare lead fiber of either 5 or
10~km, placed inside a sound-proof box so that it is isolated from ambient vibration,
followed by a 100~m armored trailing section. Controlled disturbance events are applied
to this armored section, and the backscattered intensity distributed with the benchmark is
collected from its first 50~m at 12 adjacent spatial positions. The pulse repetition rate,
which sets the temporal sampling rate of each position, is 12.5~kHz with the 5-km and
8~kHz with the 10-km lead fiber~\cite{cao2023}.

Each raw sample is a multi-channel spatio-temporal matrix. A single sample
comprises $12$ spatial positions within that 50~m span and $T = 10000$ temporal
measurements per position, so the $i$-th raw sample is written as
\begin{equation}
\label{eq:raw}
X_i = \big[\,x_{i,\ell}[t]\,\big] \in \mathbb{R}^{12 \times T},
\qquad T = 10000;\;
\ell = 1,\dots,12;\; t = 0,\dots,T-1 ;
\end{equation}
where $x_{i,\ell}[t]$ denotes the Rayleigh-backscattered intensity at the $\ell$-th
spatial position and $t$-th temporal index of sample $i$, and the associated class
label is $y_i \in \{0,1,2,3,4,5\}$; the class index $c$ used from
Section~\ref{sec:fewshot} onwards runs over the same six values. The six operationally relevant categories are
Background, Digging, Knocking, Watering, Shaking, and Walking: Background is the
baseline non-event condition, and the remaining five are intentional physical
disturbances applied to the sensing fiber.
 
For the experiments reported here, the original recordings are reduced to a
class-balanced image dataset. Each class contributes 512 images to a train-side pool and
120 images to a fixed held-out test set, giving 632 images per class and 3,792 images in
total; Table~\ref{tab:distribution} summarizes the resulting class distribution.

The subset is obtained by a single deterministic downsampling step applied to the raw
recordings before any image is generated. The train and test partitions distributed with
the benchmark are processed \emph{separately}, so that no test recording can enter the
train-side pool. Within each partition, 512 train-side and 120 test recordings per class
are drawn without replacement with a fixed seed.

This selection is performed once, at image-generation time, and is independent of the
support sampling described below. The split is therefore identical in every run, and only
the support draw varies; the standard deviations reported in Section~\ref{sec:experiments}
consequently reflect the variability of the support draw, together with the training
randomness tied to the same seed. Support sets are drawn exclusively from the train-side pool; with each support draw, 20\% of the remaining train-side images per class (at most 120; 77 per class at $K=128$) are drawn from the same pool as a validation subset for model selection. The held-out test set is not accessed during support selection, backbone screening, or tuning, which controls both selection- and index-level leakage.

Because the benchmark provides no acquisition-session identifiers, the split can only be
enforced at the image level; the resulting lack of recording-session independence is
discussed as a limitation in Section~\ref{sec:conclusion}.
 
\begin{table}[H]
\centering
\captionsetup{skip=3pt}
\caption{Event distribution of the publicly distributed version of the benchmark used in this
study (train and test partitions combined: 12,335 train and 3,084 test recordings) and the
balanced subset drawn from it. Cao \textit{et al.}~\cite{cao2023} report 15,612 recordings; the
distributed version contains 15,419, with fewer Background (2,946 vs 3,094) and Watering
(2,253 vs 2,298) recordings. The train-side pool is drawn from the benchmark's train partition
and the held-out test set from its test partition.}
\label{tab:distribution}

\scriptsize
\renewcommand{\arraystretch}{0.75}
\setlength{\tabcolsep}{3pt}

\begin{tabular}{lccccc}
\toprule
Class & Label & Benchmark recordings & Train-side pool & Held-out test & Subset total \\
\midrule
Background & 0 & 2,946 & 512 & 120 & 632 \\
Digging    & 1 & 2,512 & 512 & 120 & 632 \\
Knocking   & 2 & 2,530 & 512 & 120 & 632 \\
Watering   & 3 & 2,253 & 512 & 120 & 632 \\
Shaking    & 4 & 2,728 & 512 & 120 & 632 \\
Walking    & 5 & 2,450 & 512 & 120 & 632 \\
\midrule
Total & -- & 15,419 & 3,072 & 720 & 3,792 \\
\bottomrule
\end{tabular}
\end{table}
 
\subsection{Image Generation from Raw $\Phi$-OTDR Signals via STFT Spectrogram Mosaics}
\label{sec:imagegen}

\begin{figure}[!htbp]
\centering
\includegraphics[width=\textwidth]{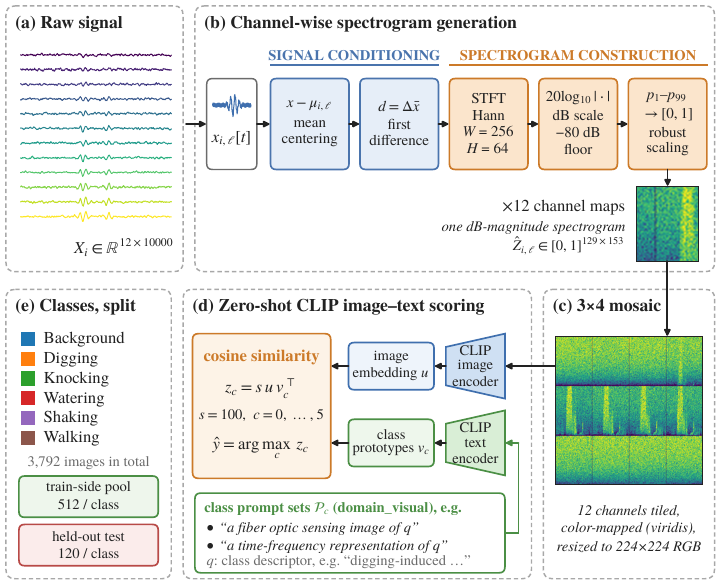}
\caption{Spectrogram-mosaic generation and zero-shot Contrastive Language--Image Pre-training (CLIP) scoring for
phase-sensitive optical time-domain reflectometry ($\Phi$-OTDR) event classification.
(a) A raw sample contains 12 channels with 10,000 measurements each.
(b) Channel-wise mean centering and first differencing precede a
one-sided Hann-window STFT ($W=256$, $H=64$, no padding).
Relative dB magnitudes from Eq.~\eqref{eq:logmag} undergo vertical
flipping and first/99th-percentile clipping and scaling to $[0,1]$.
(c) The 12 maps are tiled into a row-major $3\times4$ mosaic with
one-pixel separators, mapped to RGB with viridis, and resized to
$224\times224$ using LANCZOS interpolation.
(d) Model-specific preprocessing and frozen CLIP image encoding
yield an $L_2$-normalized embedding $u$. Prompt embeddings from
the frozen text encoder are individually normalized, averaged
within each class, and renormalized to form the rows of $\mathbf{V}$.
The logits $z_{\mathrm{text}}=100uV^{\top}$ determine the zero-shot
prediction by argmax and also feed the fusion stage of Tip-Adapter (Section~\ref{sec:fewshot}).
(e) Six classes with 512 train-side images for support/validation
and 120 held-out test images per class (3,792 images in total).
The traces and spectrograms illustrate the transformation workflow.}
\label{fig:pipeline}
\end{figure}

The goal of the image-generation pipeline is to convert each raw
$12 \times 10000$ \Photdr{} matrix into a CLIP-compatible RGB image while
preserving two types of information critical for event recognition: the
intra-position temporal--frequency signatures and the inter-position sequential
context across the 12 fiber channels. To this end, our representation employs
channel-wise STFT dB-magnitude spectrograms
arranged as a sequentially ordered $3 \times 4$ mosaic. The complete workflow is shown
in Fig.~\ref{fig:pipeline}, and the exact implementation parameters used to generate
the images consumed by the CLIP encoder are consolidated in the companion code package to be released upon publication.
 
\subsubsection{Channel-wise preprocessing}
Each raw trace $x_{i,\ell}[t]$ is mean-centered by its temporal mean $\mu_{i,\ell}$
(removing the static per-channel offset) and then temporally differenced to accentuate
dynamic, disturbance-induced variations:
\begin{equation}
\label{eq:mean}
\bar{x}_{i,\ell}[t] = x_{i,\ell}[t] - \mu_{i,\ell},
\qquad \mu_{i,\ell} = \frac{1}{T}\sum_{t'=0}^{T-1} x_{i,\ell}[t'] ;
\end{equation}
\begin{equation}
\label{eq:diff}
d_{i,\ell}[t] = \bar{x}_{i,\ell}[t+1] - \bar{x}_{i,\ell}[t] ,
\qquad t = 0,\dots,T-2 .
\end{equation}
The differenced signal $d_{i,\ell}[t]$, of length $T-1$, feeds the time--frequency transform; this
preprocessing is lightweight, deterministic, and free of trainable operations. Mean
centering is retained for consistency with the reference implementation; note that the
first difference in Eq.~\eqref{eq:diff} already removes any constant per-channel offset,
so Eq.~\eqref{eq:mean} does not by itself alter the spectrogram.
 
\subsubsection{STFT dB-magnitude spectrogram extraction}
Each differenced trace is analyzed with a Hann window $w[n]$ of length $W=256$ and hop
$H=64$ (no padding, normalized $f_s=1.0$), yielding a one-sided spectrogram of $129$
frequency bins $k=0,\dots,128$ $\times$ $153$ frames $m=0,\dots,152$ per channel; since
the benchmark pools recordings acquired at two pulse repetition rates
(Section~\ref{sec:dataset}) and the same transform is applied to all of them, the
frequency axis is expressed in normalized cycles/sample. The complex STFT coefficient is
\begin{equation}
\label{eq:stft}
Y_{i,\ell}[m,k] = \sum_{n=0}^{W-1} d_{i,\ell}[\,n + mH\,]\, w[n]\,
\exp\!\left(-\,j\,\frac{2\pi k n}{W}\right) ;
\end{equation}
where $j$ is the imaginary unit. To compress dynamic range and reveal weaker
transient structures alongside strong peaks, a decibel (dB) magnitude compression
with an $80$~dB floor is applied rather than a simple $\log(1+|Y|)$ transform.
Throughout this section, $\mathrm{clip}(\cdot,a,b)$ denotes element-wise clamping of its
argument to the interval $[a,b]$; it is a numerical operation and is unrelated to the
CLIP model.
Writing $\lvert Y_{i,\ell}\rvert$ for the STFT magnitude and $\max\lvert Y_{i,\ell}\rvert$
for its maximum over all $(m,k)$,
\begin{equation}
\label{eq:logmag}
Z_{i,\ell}[m,k] = \mathrm{clip}\!\left(20\,\log_{10}\!\left(\frac{\lvert Y_{i,\ell}[m,k]\rvert}{\max\lvert Y_{i,\ell}\rvert + \rho} + \rho\right),\; -80,\; 0\right) ;
\end{equation}
where $\rho = 10^{-8}$ both stabilizes the per-channel reference $\max\lvert Y_{i,\ell}\rvert$ and
guards the logarithm against zero, exactly as in our implementation; each bin is
expressed in dB relative to the per-channel maximum and floored at $-80$~dB by the clip
($\texttt{top\_db}=80$), so silent bins map to $-80$~dB. The matrix $Z_{i,\ell}$ is the
two-dimensional time--frequency representation of channel $\ell$, and this extraction is
performed independently for all 12 spatial positions. Because the reproducible
pipeline operates on this dB-scaled magnitude, the representation is referred to as a
dB-magnitude spectrogram throughout.
 
\subsubsection{Robust normalization, $3\times4$ spatial tiling, and RGB mapping}
Each dB spectrogram is first flipped vertically for image display and then normalized
to a bounded intensity range via robust percentile scaling, using its 1st and 99th
percentiles $p_1$ and $p_{99}$ rather than the raw minimum and maximum,
\begin{equation}
\label{eq:norm}
\hat{Z}_{i,\ell} =
\mathrm{clip}\!\left(\frac{\mathrm{clip}\!\left(Z_{i,\ell},\, p_1,\, p_{99}\right) - p_1}{p_{99} - p_1 + \rho},\; 0,\; 1\right) ;
\end{equation}
where $\rho$ is the guard constant of Eq.~\eqref{eq:logmag} and
$\hat{Z}_{i,\ell} \in [0,1]^{129 \times 153}$; percentile clipping suppresses extreme
outlier bins and stabilizes the visual dynamic range across channels and events. The 12
normalized maps are then tiled, in their original sequential channel order, into a compact
$3 \times 4$ grid separated by one-pixel borders of value zero:
\begin{equation}
\label{eq:mosaic}
M_i =
\begin{bmatrix}
\hat{Z}_{i,1}  & \hat{Z}_{i,2}  & \hat{Z}_{i,3}  & \hat{Z}_{i,4}  \\
\hat{Z}_{i,5}  & \hat{Z}_{i,6}  & \hat{Z}_{i,7}  & \hat{Z}_{i,8}  \\
\hat{Z}_{i,9}  & \hat{Z}_{i,10} & \hat{Z}_{i,11} & \hat{Z}_{i,12}
\end{bmatrix} \in [0,1]^{389 \times 615},
\end{equation}
where $389 = 3 \cdot 129 + 2$ and $615 = 4 \cdot 153 + 3$. The mosaic is mapped to color
through the deterministic \texttt{viridis} colormap $\Psi(\cdot)$, so that the separators
take the darkest colormap value $\Psi(0)$, and is then resized by LANCZOS interpolation to
the input resolution of the CLIP image encoder, without preserving its aspect ratio, and
stored as an 8-bit RGB PNG:
\begin{equation}
\label{eq:resize}
I_i = \mathrm{Resize}_{\mathrm{LANCZOS}}\!\left(\Psi(M_i),\, 224 \times 224\right),
\qquad I_i \in \{0,\dots,255\}^{224 \times 224 \times 3}.
\end{equation}
At inference and cache construction, each stored image is finally passed through the
model-specific \texttt{open\_clip} preprocessing transform before entering the CLIP
encoder (full parameters in the companion code package to be released upon publication).
 
This transformation converts a high-dimensional, multi-channel time-series sample into a
compact, structurally rich visual object. Unlike isolated
single-channel spectrogram classification, the mosaic retains cross-position
spatial context, allowing the vision encoder to assess whether event energy is
localized, repetitive, smooth, impulsive, or spatially coherent across neighboring
fiber segments. Because the dB compression of Eq.~\eqref{eq:logmag} and the percentile
scaling of Eq.~\eqref{eq:norm} are both applied per channel, absolute inter-channel
amplitude differences are not preserved; the mosaic therefore encodes the
time--frequency texture of each position and how that texture varies along the fiber,
rather than the absolute spatial energy profile. A sample-level normalization variant
that retains inter-channel amplitude is left for future ablation.
It should be noted, further, that the $3 \times 4$ layout preserves
the \emph{global sequential order} of the 12 sensing positions but does not exactly
preserve every one-dimensional fiber-neighbor adjacency: channels that are contiguous
along the fiber (e.g., positions 4 and 5) are placed on different rows, while the
layout also introduces artificial vertical adjacencies (e.g., positions 1 and 5).
We therefore interpret the mosaic as a compact ordered visual encoding rather than an
exact physical-topology map, and a direct comparison against $1 \times 12$ strip,
$4 \times 3$, and shuffled-channel layouts is left for future ablation. In summary,
each raw \Photdr{} sample is turned into a 12-channel sequence of STFT dB-magnitude
spectrograms, tiled into a $3 \times 4$ mosaic, color-mapped to RGB, resized to $224 \times 224$, and used as the visual input to the CLIP image encoder.

\subsubsection{Representative per-class spectrogram mosaics}
\begin{figure}[htbp]
\centering
\includegraphics[width=\textwidth]{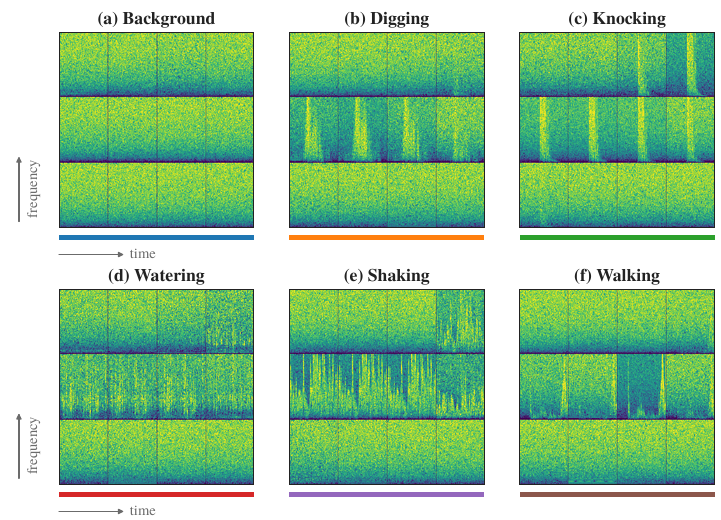}
\caption{Representative short-time Fourier transform (STFT) dB-magnitude spectrogram-mosaic examples for the six
event classes, drawn from the balanced dataset. Each panel is one $3\times4$ mosaic
(12 fiber channels) for a single class. The class-name headers are added for illustration
only and are not part of the $224\times224$ images given to the Contrastive Language--Image Pre-training (CLIP) image encoder. The panels are the unmodified $224\times224$ mosaics.}
\label{fig:classexamples}
\end{figure}

Figure~\ref{fig:classexamples} shows one representative spectrogram-mosaic example per
event class, drawn from the balanced dataset. The classes exhibit
visually distinct time--frequency textures: Background
is a diffuse, structure-less field; Digging produces wide, high-energy vertical bursts;
Knocking appears as short, sharp impulsive stripes; Watering is a smoother, sustained,
distributed texture; Shaking shows continuous, horizontally coherent oscillatory bands;
and Walking forms a repetitive sequence of evenly spaced footstep spikes. The physical
similarity between the sustained textures of Shaking and Watering foreshadows the
dominant Shaking--Watering confusion analyzed in Section~\ref{sec:experiments}.
 
\subsection{Few-Shot Vision-Language Model}
\label{sec:fewshot}

Few-shot VLM classification is formulated as an episodic
adaptation problem over the spectrogram-mosaic representation of our dataset. Each
raw \Photdr{} recording of the balanced subset is converted into a CLIP-compatible RGB mosaic
(Section~\ref{sec:imagegen}); the train-side pool and the fixed held-out test set are
therefore the images of the recordings already assigned to them before image generation
(Section~\ref{sec:dataset}). For each class, 512 images are available for
support and validation sampling and 120 are reserved for final testing; crucially,
the test set is never accessed during prompt selection, backbone screening,
hyperparameter optimization, or support-set construction.
 
For a six-class task with $K$ labeled examples per class, a few-shot
episode is defined by a support set
$S_K = \{(I_i, y_i)\}_{i=1}^{6K}$, where $I_i$ is a spectrogram-mosaic image
(Eq.~\eqref{eq:resize}) and $y_i \in \{0,\dots,5\}$ its event label. Candidate shot values
$K \in \{1,2,4,8,16,32,64,128\}$ are evaluated systematically to quantify how
performance improves as the number of annotated examples grows. Rather than
training the VLM backbone end-to-end, the pre-trained CLIP image and text encoders provide a
fixed representation space, and adaptation is carried out through a lightweight
support-set cache; the complete architecture, with fixed and trained components made
explicit, is shown in Fig.~\ref{fig:tipf}.
 
\subsubsection{CLIP}
CLIP comprises an image encoder $f_I(\cdot)$ and a text encoder $f_T(\cdot)$ that
are jointly trained to align visual and textual representations in a shared
embedding space of dimension $C$~\cite{radford2021clip}. In our setting, a spectrogram mosaic $I$ is mapped to an $L_2$-normalized image embedding $u \in \mathbb{R}^{1 \times C}$, and each
event class $c \in \{0,\dots,5\}$ is described by a set $\mathcal{P}_c$ of textual
prompts, whose individually $L_2$-normalized text embeddings are averaged and
renormalized into a single class prototype $v_c \in \mathbb{R}^{1 \times C}$. All embeddings in this study are treated as row vectors, so an inner product between two of them is written with the second factor transposed. Zero-shot classification is then obtained from the scaled cosine similarities between $u$ and each $v_c$:
\begin{equation}
\label{eq:clip}
u = \frac{f_I(I)}{\lVert f_I(I)\rVert_2}, \qquad
v_c = \frac{\bar{v}_c}{\lVert \bar{v}_c\rVert_2}, \quad
\bar{v}_c = \frac{1}{|\mathcal{P}_c|}\sum_{\pi \in \mathcal{P}_c}
\frac{f_T(\pi)}{\lVert f_T(\pi)\rVert_2} ;
\end{equation}
\begin{equation}
\label{eq:zs}
z_c = s\, u\, v_c^{\top}, \qquad
z_{\mathrm{text}}(u) = [z_0,\dots,z_5] = s\, u\, \mathbf{V}^{\top}, \qquad
\hat{y} = \arg\max_{c} z_c ;
\end{equation}
where the rows of $\mathbf{V} \in \mathbb{R}^{6 \times C}$ are the prototypes $v_0,\dots,v_5$, and $s$ is the logit-scale factor that rescales the cosine similarities, which lie in
$[-1,1]$, to the range a softmax over class scores expects. It is held at the constant
$s=100$ for every backbone rather than read from each checkpoint's own learned
\texttt{logit\_scale}, so that the text branch is scaled identically across the OpenAI and
LAION encoders compared in Section~\ref{sec:backbone}, and it is never updated.
This zero-shot pathway is a diagnostic
baseline that tests whether natural-language class descriptions alone can classify
\Photdr{} event images. In our experiments zero-shot CLIP stays close to chance,
showing that direct language--image alignment is inadequate for this specialized
domain; nevertheless, the pre-trained image-embedding space retains substantial
structure that parameter-efficient few-shot adaptation can exploit. The zero-shot
scoring pathway is retained as the text-logit branch of the Tip-Adapter variants.
 
\subsubsection{Tip-Adapter}
Tip-Adapter augments the zero-shot CLIP classifier with a training-free key--value
cache built directly from the $K$-shot support set~\cite{zhang2022tipadapter}. The
cache keys $\mathbf{F}_{\mathrm{train}} \in \mathbb{R}^{6K \times C}$ are the
$L_2$-normalized CLIP image embeddings of the $6K$ support images, and the cache values
$\mathbf{L}_{\mathrm{train}} \in \mathbb{R}^{6K \times 6}$ are their one-hot labels. For a
query embedding $u$ obtained exactly as in Eq.~\eqref{eq:clip}, the affinity between $u$
and each cache key is computed with an exponential kernel modulated by a sharpness
parameter $\beta$:
\begin{equation}
\label{eq:affinity}
A(u)=\exp\!\left[-\beta\left(1-u\,\mathbf{F}_{\mathrm{train}}^{\top}\right)\right],
\qquad
z_{\mathrm{cache}}(u)=A(u)\,\mathbf{L}_{\mathrm{train}} .
\end{equation}
The Tip-Adapter logits fuse these retrieved cache logits with the CLIP text logits
through a residual weighting coefficient $\alpha$:
\begin{equation}
\label{eq:fusion}
z(u) = z_{\mathrm{text}}(u) + \alpha\, z_{\mathrm{cache}}(u) .
\end{equation}
Here $z_{\mathrm{text}}(u) \in \mathbb{R}^{1 \times 6}$ collects the six scaled
image--text similarities $z_c$ of Eq.~\eqref{eq:zs}, that is, the zero-shot evidence
supplied by the language branch, whereas $z_{\mathrm{cache}}(u) \in \mathbb{R}^{1 \times 6}$
is the affinity-weighted vote of the labeled support images, that is, the evidence
supplied by visual similarity to the cache. Their sum $z(u)$ is the six-class score from
which the predicted label is taken.
Because Eqs.~\eqref{eq:affinity}--\eqref{eq:fusion} involve no gradient-based
optimization, Tip-Adapter is a computationally efficient and interpretable
baseline. For the training-free Tip-Adapter, the fusion weight $\alpha$ and the
sharpness $\beta$ are selected for each support draw by validation accuracy over
$\alpha \in \{0.5,1,2,3,5,7\}$ and $\beta \in \{1,3,5,7,10,15\}$, following the
original method~\cite{zhang2022tipadapter}. For Tip-Adapter-F, both are fixed at $\alpha = 0.5$ and $\beta = 3.0$, in training and at inference, throughout the Stage-2 search and all held-out evaluations (Section~\ref{sec:expdetails}).
 
\subsubsection{Tip-Adapter-F}
\begin{figure}[htbp]
\centering
\includegraphics[width=\textwidth]{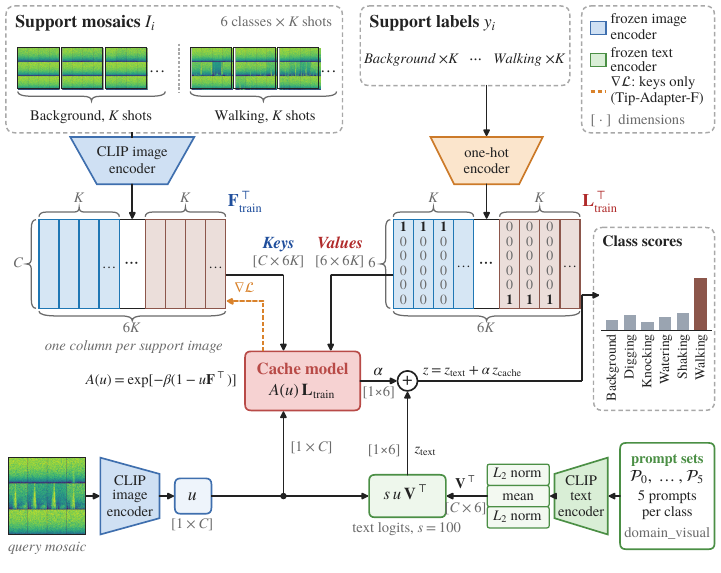}
\caption{Tip-Adapter and Tip-Adapter-F architecture. Top: the $6K$ support mosaics and their labels are encoded by the frozen Contrastive Language--Image Pre-training (CLIP) image encoder and a one-hot encoder. Middle: the cache keys $\mathbf{F}_{\mathrm{train}}^{\top}$ ($C\times6K$) and values
$\mathbf{L}_{\mathrm{train}}^{\top}$ ($6\times6K$) are shown with one column vector per support image, where $K$ is the number of shots per class and $C$ the embedding dimension. Bottom: a query mosaic is encoded into $u$ ($1\times C$), which yields the cache logits $z_{\mathrm{cache}}=A(u)\,\mathbf{L}_{\mathrm{train}}$ with
$A(u)=\exp[-\beta(1-u\mathbf{F}^{\top})]$ (element-wise) and the text logits
$z_{\mathrm{text}}=s\,u\mathbf{V}^{\top}$ ($s=100$); the rows of $\mathbf{V}$ are the class prototypes obtained by normalizing, averaging and renormalizing the prompt embeddings of the frozen text encoder. The fused logits $z=z_{\mathrm{text}}+\alpha z_{\mathrm{cache}}$ give the class scores ($\alpha=0.5$, $\beta=3.0$ for Tip-Adapter-F; validation-selected for Tip-Adapter). Tip-Adapter uses $\mathbf{F}=\mathbf{F}_{\mathrm{train}}$ without training;
Tip-Adapter-F initializes $\mathbf{F}'_{\mathrm{train}}$ from $\mathbf{F}_{\mathrm{train}}$, updates only these keys with the support-set cross-entropy (dashed arrow), and uses
$\mathbf{F}=\mathbf{F}'_{\mathrm{train}}$ at inference. Both CLIP encoders and the one-hot values remain fixed; the class scores shown are illustrative.}
\label{fig:tipf}
\end{figure}

Tip-Adapter-F extends Tip-Adapter by treating the support cache keys as
\emph{trainable} parameters, while the CLIP image encoder, the CLIP text encoder,
and the cache values $\mathbf{L}_{\mathrm{train}}$ remain fixed~\cite{zhang2022tipadapter}.
The keys are initialized from the support features and then optimized over a few
epochs by minimizing the cross-entropy of the fused logits of Eq.~\eqref{eq:fusion}:
\begin{equation}
\label{eq:tipf}
\mathcal{L}=\mathrm{CE}\!\left(y,\; z\!\left(u;\mathbf{F}'_{\mathrm{train}}\right)\right) ;
\end{equation}
where $\mathbf{F}'_{\mathrm{train}}$ denotes the learnable cache keys, updated by gradient
descent while the gradient never reaches the frozen encoders. Restricting the learnable parameters to
the cache alone gives Tip-Adapter-F markedly higher parameter efficiency than full
fine-tuning of a deep network. Figure~\ref{fig:tipf} details the complete architecture
and makes explicit which components are frozen and which are trained; the validation
protocol and selected configuration are detailed in Section~\ref{sec:expdetails}.
 
\bigskip

\subsubsection{Optimization protocol and parameter efficiency}
Only the cache keys are trainable; the CLIP encoders and the one-hot cache values
$\mathbf{L}_{\mathrm{train}}$ stay fixed, and gradient batches are drawn exclusively from the support set. Validation images never contribute gradients: they are used only for model selection, namely the backbone ranking and configuration search of Section~\ref{sec:expdetails} and the per-draw choice of $\alpha$ and $\beta$ for the training-free Tip-Adapter, and since no early stopping is applied they play no part in selecting the epoch at which training stops. At 128 shots the cache holds $6\times128=768$ keys, each of dimension $C=768$ for the selected ViT-L-14/openai backbone, that is, $589{,}824$ trainable scalars, orders of magnitude fewer than end-to-end backbone fine-tuning. Keys are initialized from the support features and optimized for 30 epochs with AdamW~\cite{loshchilov2019adamw} (lr $0.002$, weight decay $10^{-4}$, non-default $\varepsilon=10^{-4}$ matching our implementation), cosine annealing~\cite{loshchilov2017sgdr} stepped once per epoch, batch size 16 (the whole support set when $6K < 16$), and mean-reduced cross-entropy on the fused logits; no augmentation, mixed precision, or early stopping is used, and all runs use fixed Python/NumPy/PyTorch~\cite{paszke2019pytorch} seeds. All experiments were run with PyTorch 2.11.0 and the OpenCLIP library~\cite{ilharco2021openclip} (package \texttt{open\_clip\_torch} 2.24.0) on a single NVIDIA A100 GPU. This configuration, together with the \texttt{domain\_visual} prompts, is selected once at $K=128$
(Section~\ref{sec:expdetails}) and is then used unchanged for every shot count and every backbone reported in Fig.~\ref{fig:scaling} and in the backbone comparison (Section~\ref{sec:backbone}); no per-shot or per-backbone re-tuning is performed.

%%%%%%%%%%%%%%%%%%%%%%%%%%%%%%%%%%%%%%%%%%%%%%%%%%%%%

\section{Experiment Details, Results, and Discussion}
\label{sec:experiments}

This section presents the final evaluation on our dataset, conducted under a
strictly test-isolated image-level protocol. The objective is to determine whether pre-trained
CLIP representations, when adapted through Tip-Adapter-F, yield a label-efficient
event classifier for the signal-derived spectrogram mosaics of
Section~\ref{sec:imagegen}. Unless stated otherwise, all reported metrics are averaged
over 20 independent support-sampling trials, that is, 20 distinct random draws of the
support set from the train-side pool.
Section~\ref{sec:expdetails} describes the two-stage selection protocol,
Section~\ref{sec:evalframework} the evaluation metrics and the backbone screening
outcome, Section~\ref{sec:prompt} the prompt-family analysis,
Section~\ref{sec:backbone} the backbone comparison, Section~\ref{sec:umap} the
embedding-space diagnostics, and Section~\ref{sec:uncertainty} the measurement uncertainty of
the reported accuracies.
 
\subsection{Experiment Details}
\label{sec:expdetails}

Model selection proceeds in two stages, both restricted entirely to the train-side pool;
the test split is accessed only after the final operating point is fixed. In Stage~2
and in every held-out evaluation of Tip-Adapter-F, the cache-fusion parameters are held
at $\alpha = 0.5$ and $\beta = 3.0$, so the Stage-2 search covers only the prompt family,
the shot count, and the cache-training hyperparameters.

In Stage~1, eight CLIP encoders are screened with Tip-Adapter-F under a fixed training
setting (lr $10^{-3}$, 20 epochs, batch size 32, and $\alpha = 1$, $\beta = 5$ in the
training loss); at inference, $\alpha$ and $\beta$ are then chosen on the validation
subset from the same grid as for the training-free Tip-Adapter. Because this choice and
the score use the same validation subset, Stage-1 values are mildly optimistic and are
used only for ranking. Each encoder is evaluated over the four prompt families and the six shot counts $\{4,8,16,32,64,128\}$, which gives 24 distinct settings, and every setting is repeated
over three support-sampling seeds, for 72 validation runs per encoder. The encoders are
ranked by their mean Tip-Adapter-F validation macro-F1 over these 72 runs, and the two
highest-ranked encoders proceed to Stage~2.

In Stage~2, these two encoders are swept over the four prompt families, the eight shot
counts $\{1,2,4,8,16,32,64,128\}$, three learning rates, three epoch counts, and three
batch sizes. This yields $1{,}728$ unique configurations, each repeated over the same
three seeds for a total of $5{,}184$ validation runs. Selection is lexicographic on
validation macro-F1, then accuracy, then lower false-negative rate (FNR) and lower expected
calibration error (ECE) as further tie-breakers. The FNR is the fraction of event recordings
(all classes except Background) that are classified as Background. The ECE is computed from
the maximum softmax probability of the fused logits: the predictions are grouped into ten
equal-width confidence bins, and the absolute differences between the mean confidence and the
accuracy in each bin are averaged, weighted by the number of predictions in the bin.

The complete Stage-1 and Stage-2 search spaces, run counts, selection rule, and the
resulting operating point are listed in Table~\ref{tab:searchspace}. That configuration
is frozen before the held-out test set is accessed for the first time, which keeps
test-set information entirely out of model selection and prevents optimistic bias; the
final results reported below are then averaged over the 20 support-sampling trials, which
captures support-sampling variance.
 
\begin{table}[t]
\centering
\caption{Two-stage, test-isolated validation protocol. Stage~1 selects the CLIP
backbone, while Stage~2 tunes the remaining hyperparameters with the Tip-Adapter-F
fusion weights fixed at $\alpha=0.5$ and $\beta=3.0$.
Model selection uses three support-sampling seeds (1000–1002) to keep the search tractable; the frozen configuration is then evaluated on the held-out test set over 20 support-sampling seeds, that is, 20 independent random draws of the support set.}
\label{tab:searchspace}

\scriptsize
\setlength{\tabcolsep}{4pt}
\renewcommand{\arraystretch}{0.88}

\begin{tabular}{lll}
\toprule
Setting
& Stage 1 (backbone screening)
& Stage 2 (hyperparameter tuning) \\
\midrule
CLIP backbones
& 8 candidates
& Top 2 from Stage~1 \\

Prompt families
& 4
& 4 \\

Shots per class
& $\{4,8,16,32,64,128\}$
& $\{1,2,4,8,16,32,64,128\}$ \\

Fusion $(\alpha,\beta)$
& validation-selected
& $(0.5,\,3.0)$, fixed \\

Training settings
& lr $10^{-3}$; 20 epochs; batch 32
& \begin{tabular}[t]{@{}l@{}}
  $\mathrm{lr}\in\{0.5,1,2\}\times10^{-3}$;\\[-1pt]
  epochs $\in\{10,20,30\}$; batch size $\in\{16,32,64\}$
  \end{tabular} \\

Seeds
& $\{1000,1001,1002\}$
& $\{1000,1001,1002\}$ \\
\midrule

\multicolumn{3}{l}{
\emph{Selection:} validation macro-F1, then accuracy, FNR, ECE; test data are never used.
} \\

\multicolumn{3}{l}{
\emph{Selected:} ViT-L-14/openai, \texttt{domain\_visual}, 128 shots/class,
$\mathrm{lr}=2\times10^{-3}$, 30 epochs, batch 16.
} \\
\bottomrule
\end{tabular}
\end{table}
 
\subsection{Experimental Design and Evaluation Framework}
\label{sec:evalframework}

Performance is reported with two standard classification metrics: overall accuracy and
macro-averaged F1 (macro-F1), which weights all six classes equally. Model selection at
Stage-2 prioritizes validation macro-F1, with the tie-breakers of Section~\ref{sec:expdetails}; the test split is never consulted. We report the three stages of the adopted CLIP-based approach with the validation-selected backbone and prompts: zero-shot CLIP scores classes by image--text similarity alone, the training-free Tip-Adapter augments this with a support-set cache, and Tip-Adapter-F fine-tunes the cache keys as described in Section~\ref{sec:fewshot}; the two cache-based stages share the same $K$-shot support draws. Because the validation-selected $\beta$ of the training-free Tip-Adapter reaches the largest grid value ($\beta = 15$) in 15 of the 20 draws, its accuracy is conservative, and the gap between the two cache-based stages should be read as an upper estimate of the gain from fine-tuning the cache keys.

The purpose of Stage-1 is to choose the CLIP backbone using validation data only, before
any hyperparameter tuning. Averaging each backbone over $72$ screening
runs (the $24$ settings formed by 4 prompt families $\times$ 6 shot counts, each repeated
over 3 support-sampling seeds), the Vision Transformers rank above the
ResNets, with ViT-L-14/openai first (Table~\ref{tab:stage1}). Because these means
deliberately aggregate untuned settings across all shot counts, they are lower than the tuned
Stage-2 operating point ($96.90\%$ validation macro-F1); only the relative \emph{ranking}
is used here, and no test-set information enters either stage.
 
\begin{table}[htbp]
\centering\footnotesize
\caption{Stage-1 backbone screening: the eight candidate CLIP encoders ranked by
\emph{validation} performance, before hyperparameter tuning and without any test-set
access. All rows are Tip-Adapter-F runs under the fixed Stage-1 setting, and each
averages the same $n=72$ screening runs, so the absolute values are intentionally below
the tuned Stage-2 operating point; only the ranking is used, and the two highest-ranked
encoders, ViT-L-14/openai and ViT-H-14/laion2b, proceed to Stage~2.}
\label{tab:stage1}
\begin{tabular}{lccc}
\toprule
Backbone & Val macro-F1 & Val accuracy & $n$ \\
\midrule
ViT-L-14/openai   & 83.14\% & 84.06\% & 72 \\
ViT-H-14/laion2b  & 82.10\% & 82.88\% & 72 \\
ViT-B-16/openai   & 78.52\% & 79.56\% & 72 \\
ViT-B-32/openai   & 77.50\% & 78.53\% & 72 \\
RN50/openai       & 75.43\% & 76.28\% & 72 \\
RN50x4/openai     & 73.57\% & 75.03\% & 72 \\
RN50x16/openai    & 70.34\% & 71.86\% & 72 \\
RN101/openai      & 69.89\% & 71.35\% & 72 \\
\bottomrule
\end{tabular}
\end{table}

On the held-out split (Table~\ref{tab:comparison}), the adopted method progresses through
three clear stages. Zero-shot CLIP is degenerate rather than random: at chance accuracy
($16.67\%$) its $4.78\%$ macro-F1 indicates collapse onto essentially one event class,
confirming that direct zero-shot image--text matching is ineffective for this
representation. Adding the training-free Tip-Adapter cache already lifts accuracy to
\pmv{86.40}{0.78}, so the CLIP feature space is discriminative even before any training;
fine-tuning the support cache with Tip-Adapter-F then raises accuracy to
\pmv{96.81}{0.42} and macro-F1 to \pmv{96.79}{0.42}.

The training-free Tip-Adapter row also helps to locate where the gain originates, although it is not a visual-only baseline. Because zero-shot CLIP collapses onto essentially one class, its text logits $z_{\mathrm{text}}$ do not supply useful class evidence; they mainly add a systematic preference for the collapsed class. Under Eq.~\eqref{eq:fusion}, the Tip-Adapter prediction is therefore governed by the kernel-weighted vote over the frozen CLIP embeddings of the support images, shifted by this text-branch bias. The accuracy of \pmv{86.40}{0.78} thus shows that the frozen visual features already separate the six classes well before any training. Since the bias can pull some predictions toward the collapsed class, this figure is more likely to understate than to overstate what the visual features provide on their own. The further rise to \pmv{96.81}{0.42} with Tip-Adapter-F comes from fitting the cache keys, which can also learn to compensate for the text-branch bias. These readings are an interpretation of Table~\ref{tab:comparison} rather than a controlled ablation of the text branch. 
 
\begin{table}[htbp]
\centering
\caption{Held-out test results across the three stages of the adopted CLIP-based approach
(six-class task), all with the validation-selected ViT-L-14/openai backbone and
\texttt{domain\_visual} prompts. Zero-shot CLIP uses no support set and is therefore
deterministic, so its row carries no $\pm$. The two cache-based rows use $K=128$ shots per
class and the same 20 support draws and are means $\pm$ standard deviation over these
trials; Tip-Adapter-F uses $\alpha=0.5$, $\beta=3.0$, whereas the training-free
Tip-Adapter selects $\alpha$ and $\beta$ on validation for each draw.}
\label{tab:comparison}
\begin{tabular}{lcc}
\toprule
Stage & Accuracy & Macro-F1 \\
\midrule
Zero-shot CLIP        & 16.67\%            & 4.78\%             \\
Tip-Adapter & \pmv{86.40}{0.78} & \pmv{86.17}{0.81} \\
\textbf{Tip-Adapter-F}& \pmvb{96.81}{0.42} & \pmvb{96.79}{0.42} \\
\bottomrule
\end{tabular}
\end{table}

\begin{figure}[htbp]
\centering
\includegraphics[width=0.62\textwidth]{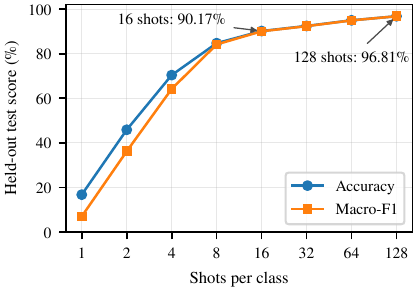}
\caption{Tip-Adapter-F few-shot scaling of accuracy and macro-averaged F1 score (macro-F1) versus support-set size (1--128 shots per class) on the fixed held-out test set, showing the 16-shot label-efficient point and the
128-shot peak.}
\label{fig:scaling}
\end{figure}

The few-shot trajectory shows two clear regimes (Fig.~\ref{fig:scaling}). With a single
support image per class, Tip-Adapter-F stays at chance level (\pmv{16.76}{0.41}); accuracy
then rises steeply up to 8 shots (\pmv{45.87}{4.55}, \pmv{70.40}{2.86} and
\pmv{84.72}{2.60} at 2, 4 and 8 shots), where the learnable cache is still sparsely
populated and the variability across support draws is largest, and flattens beyond
16 shots as the support set grows. By 16 shots, that is, only 96 labeled support
images at adaptation time across the six classes, Tip-Adapter-F reaches \pmv{90.17}{0.81} accuracy with the configuration selected at $K=128$ and not re-tuned, which we take as the practical low-label operating point and the headline label-efficiency result, reached with a small fraction of the labels required by the full-data training of the prior CLIP-based study~\cite{lin2025clipotdr}. A support size of 64 shots gives a strong performance--cost compromise, and the 128-shot configuration delivers the peak headline accuracy.
 
\begin{figure}[htbp]
\centering
\begin{subfigure}{0.5\textwidth}
\includegraphics[width=\linewidth]{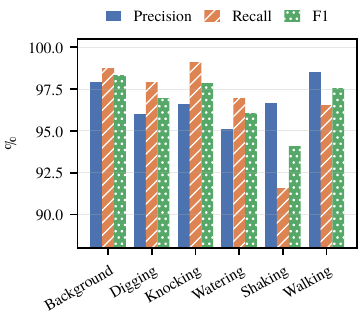}
\caption{}\label{fig:classwise}
\end{subfigure}\hfill
\begin{subfigure}{0.48\textwidth}
\includegraphics[width=\linewidth]{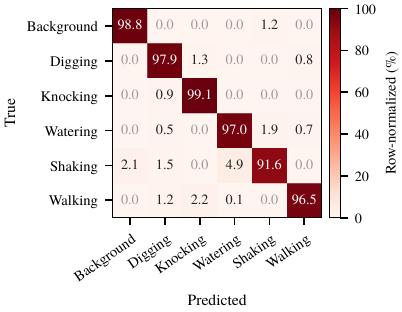}
\caption{}\label{fig:confusion}
\end{subfigure}
\caption{Class-level diagnostics for the selected Tip-Adapter-F model (Vision Transformer ViT-L-14 with OpenAI weights,
\texttt{domain\_visual}, $K=128$), computed from the held-out predictions of all 20
support-sampling trials pooled ($20 \times 120$ test images per class).
(a)~Per-class precision, recall, and F1; Shaking has the lowest recall.
(b)~Row-normalized confusion matrix; the dominant residual confusion is
Shaking$\rightarrow$Watering. Rows are normalized to sum to $100\%$ and are displayed to
one decimal place, so an individual row may not add up to exactly $100\%$.}
\label{fig:classdiag}
\end{figure}

For a given class, recall is the fraction of that class's test images the model assigns to
it, precision is the fraction of the images assigned to it that truly belong to it, and F1
is their harmonic mean. A class with low recall is therefore one whose events are
frequently absorbed by some other class, whereas low precision marks a class that absorbs
events from elsewhere.
Residual errors are unevenly distributed across the six classes: Shaking is the hardest
class ($91.58\%$ recall, $94.07\%$ F1; Fig.~\ref{fig:classwise}), and the dominant
misclassification is Shaking$\rightarrow$Watering (Fig.~\ref{fig:confusion}). Correspondingly, Watering has the lowest precision ($95.10\%$), as it is the class that
absorbs most of the misassigned Shaking events. This matches
physical intuition, since sustained oscillatory impacts and continuous flow-like
disturbances produce similar textures in the time--frequency mosaic; all other classes are
recognized with recall above $95\%$.
 
\subsection{Prompt Engineering and Model Performance}
\label{sec:prompt}

Because CLIP scores classes through image--text similarity, prompt wording can
influence prediction. We evaluate four prompt families, basic\_clean, domain\_visual
(validation-selected), confusion\_focused, and shaking\_focused, which differ in the
number and specificity of the per-class prompts (Table~\ref{tab:prompts}). Three families
apply four fiber-specific templates, ``a fiber optic sensing image of $q$'', ``a
distributed acoustic sensing pattern of $q$'', ``a processed phi-OTDR signal image of
$q$'' and ``a time-frequency representation of $q$'', to a family-specific class
descriptor $q$ and add one or two untemplated descriptions, giving five or six prompts per
class; basic\_clean uses three simpler prompts built from the same short descriptors as
domain\_visual. The full prompt lists are given in the companion code package to be
released upon publication. All prompts are tokenized with the CLIP byte-pair tokenizer,
whose context length is 77 tokens including the start and end tokens; the longest prompt
used here has 58 tokens, so no prompt is truncated. Each family defines the prompt sets
$\mathcal{P}_c$ of Eq.~\eqref{eq:clip}, so prompt wording enters the model only through
the class prototypes $v_c$, that is, through the text logits $z_{\mathrm{text}}(u)$ of
Eq.~\eqref{eq:zs} that are fused with the support-cache logits through
Eqs.~\eqref{eq:affinity}--\eqref{eq:fusion}.

\begin{table}[t]
\centering
\caption{The four evaluated prompt families. The four fiber-specific templates are
listed in Section~\ref{sec:prompt}; $q$ denotes a per-class descriptor.}
\label{tab:prompts}

\footnotesize
\setlength{\tabcolsep}{4pt}
\renewcommand{\arraystretch}{0.95}

\begin{tabularx}{\linewidth}{
    >{\raggedright\arraybackslash}p{0.25\linewidth}
    >{\raggedright\arraybackslash}X}
\toprule
Prompt family & Representative content \\
\midrule

\texttt{basic\_clean}
& Three prompts per class built from a short descriptor $q$: $q$ itself,
``a clean spectrogram image of $q$'' and ``a fiber optic sensing image showing $q$''. \\

\texttt{domain\_visual} (selected)
& The four templates applied to the short descriptor $q$ (e.g.\ ``digging-induced
vibration pattern in a fiber optic sensing image''), plus a descriptive per-class
paragraph; five prompts per class. \\

\texttt{confusion\_focused}
& The four templates applied to contrastive descriptors that state what each class is
not (e.g.\ ``dig class: irregular excavation activity, not a single knock, not smooth
water, not periodic shaking, not clean footsteps''), plus the descriptor itself; five
prompts per class. \\

\texttt{shaking\_focused}
& The four templates applied to descriptors written to separate Shaking from Watering
and Walking, plus that descriptor and the \texttt{confusion\_focused} descriptor; six
prompts per class. \\

\bottomrule
\end{tabularx}
\end{table}

A post-selection prompt-family ablation on the held-out set shows that the effect of
prompt choice is marginal (Table~\ref{tab:promptabl}): basic\_clean is numerically
highest, the validation-selected domain\_visual follows $0.01$~pp behind, and
confusion\_focused and shaking\_focused are a further $0.07$~pp lower, giving a spread of
about $0.08$~pp across the four families. The four families are therefore numerically very
similar relative to the observed across-seed variability of the selected configuration
($0.42$~pp); no formal significance test was performed, and the ranking between them
should not be read as a preference. domain\_visual is retained as the validation-selected family, since
switching to basic\_clean on test accuracy alone would amount to test-set selection. At 128 shots, performance is therefore insensitive to the precise prompt wording, even for the simpler basic\_clean prompts: the learned cache governs the decision.

\begin{table}[htbp]
\centering
\caption{Post-selection prompt-family ablation on the held-out test set, at the selected
operating point (ViT-L-14/openai, 128 shots). The spread across families is smaller than
the across-seed standard deviation of the selected configuration; values are means $\pm$ standard deviation over the same 20 support draws, and no formal significance
test was performed.}
\label{tab:promptabl}
\footnotesize
\begin{tabular}{lcc}
\toprule
Prompt family & Test accuracy & Test macro-F1 \\
\midrule
\texttt{basic\_clean}                    & \pmv{96.82}{0.43} & \pmv{96.81}{0.44} \\
\texttt{domain\_visual} (selected)       & \pmv{96.81}{0.42} & \pmv{96.79}{0.42} \\
\texttt{confusion\_focused}              & \pmv{96.74}{0.43} & \pmv{96.73}{0.43} \\
\texttt{shaking\_focused}                & \pmv{96.74}{0.37} & \pmv{96.72}{0.37} \\
\bottomrule
\end{tabular}
\end{table}
% TODO (Ahmet): add the seed-level standard deviations for confusion_focused and
% shaking_focused, and if available a macro-F1 column for all four families.

\subsection{Tip-Adapter-F Backbone Comparison}
\label{sec:backbone}

\begin{figure}[htbp]
\centering
\begin{subfigure}{0.5\textwidth}
\includegraphics[width=\linewidth]{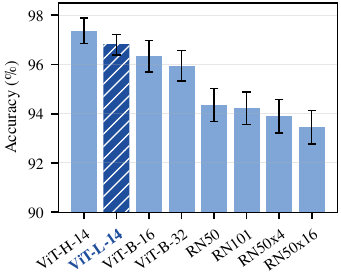}
\caption{}\label{fig:bbabl}
\end{subfigure}\hfill
\begin{subfigure}{0.48\textwidth}
\includegraphics[width=\linewidth]{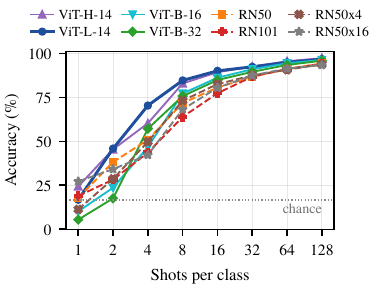}
\caption{}\label{fig:bbshot}
\end{subfigure}
\caption{Tip-Adapter-F comparison of Contrastive Language--Image Pre-training (CLIP) image-encoder backbones (ViT: Vision Transformer; RN: ResNet) on the held-out test set; no Stage-1 or
Stage-2 validation numbers appear in either panel. (a)~Backbone ablation at the 128-shot operating point (means $\pm$ standard deviation over 20 support draws); ViT-H-14 attains the highest test accuracy, but the validation-selected ViT-L-14/openai (hatched bar) is kept to avoid test-set bias. (b)~Backbone$\times$shot accuracy trajectories with the same configuration; each backbone has its own marker, ResNets are dashed, and the dotted line marks chance level; from 8 shots onwards Vision Transformers lead ResNets, with gaps narrowing as support grows.}
\label{fig:backbone}
\end{figure}

We evaluate ResNet~\cite{he2016resnet} (RN50, RN101, RN50x4, RN50x16) and Vision
Transformer~\cite{dosovitskiy2021vit} (ViT-B-32, ViT-B-16, ViT-L-14, ViT-H-14) CLIP
encoders via OpenCLIP~\cite{ilharco2021openclip} with OpenAI and LAION-2B (the English subset of LAION-5B~\cite{schuhmann2022laion5b}) weights. In every case both CLIP encoders are kept
fixed and only the cache keys are trained, so the comparison isolates the quality of the
pre-trained visual representation. Both panels of Fig.~\ref{fig:backbone} report
Tip-Adapter-F accuracy on the held-out test set.

At the 128-shot operating point the Vision Transformers outperform the ResNets without
exception (Fig.~\ref{fig:bbabl}). From 8 shots onwards the same ordering holds, with the
gap narrowing as the support set grows (Fig.~\ref{fig:bbshot}); below 8 shots the curves
cross and lie near or below chance at one shot, so no backbone ranking is meaningful in
that regime. The test ordering broadly reproduces the Stage-1 validation ranking of Table~\ref{tab:stage1}: the two largest Vision Transformers lead and the ResNets trail in both, although the top two exchange places on test. The protocol selects ViT-L-14/openai, and a post-hoc test-side comparison finds ViT-H-14/laion2b higher (\pmv{97.38}{0.52}). The validation-selected
model is nevertheless kept, because adopting ViT-H-14 on the strength of a test-set number
would be exactly the selection bias the protocol is designed to avoid.
 
\subsection{Embedding-Space Analysis}
\label{sec:umap}

\begin{figure}[htbp]
\centering
\begin{subfigure}{0.49\textwidth}
\includegraphics[width=\linewidth]{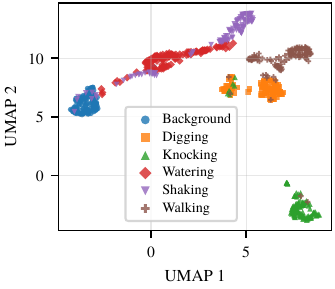}
\caption{}\label{fig:umapclass}
\end{subfigure}\hfill
\begin{subfigure}{0.49\textwidth}
\includegraphics[width=\linewidth]{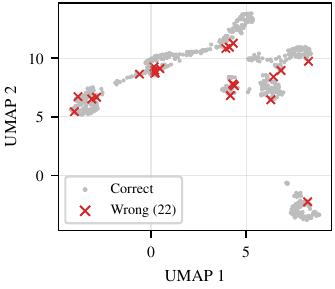}
\caption{}\label{fig:umapcorr}
\end{subfigure}
\caption{Uniform manifold approximation and projection (UMAP) of the frozen Vision Transformer ViT-L-14 (OpenAI weights) image embeddings of the held-out test mosaics. (a)~Colored and marked by true class: classes form largely separated clusters, with
Shaking/Watering adjacent. (b)~Marked by Tip-Adapter-F prediction correctness in the first support-sampling trial (22 errors): errors (crosses) are frequent near cluster boundaries.}
\label{fig:umap}
\end{figure}

To interpret where the residual errors originate, we project the frozen ViT-L-14/openai image embeddings of the 720 held-out test mosaics into two dimensions with UMAP~\cite{mcinnes2018umap} ($n_{\text{neighbors}}=15$, $\text{min\_dist}=0.1$) (Fig.~\ref{fig:umap}). Colored by true class (Fig.~\ref{fig:umapclass}), most event classes form largely separated clusters, whereas Shaking and Watering, the dominant confusion pair, lie adjacent with partial overlap, consistent with the confusion matrix (Fig.~\ref{fig:confusion}). Colored by the correctness of the selected Tip-Adapter-F model in the first support-sampling trial (Fig.~\ref{fig:umapcorr}), many of the misclassifications lie at or near these inter-cluster boundaries, supporting the interpretation that the residual errors form a small set of physically ambiguous, decision-boundary cases. Whether a confidence-based triage rule can actually isolate them is not evaluated here and is left for future work.
 
\FloatBarrier
\subsection{Measurement Uncertainty}
\label{sec:uncertainty}
Two independent sources of uncertainty affect every accuracy reported above
(table~\ref{tab:uncertainty}). The first is the choice of the $K$-shot support
set, which we quantify by repeating each evaluation over 20 support draws while
keeping the held-out test set fixed. For the selected Tip-Adapter-F configuration
the accuracy over these draws is $96.81\pm0.42\%$ (range $95.97$--$97.36\%$), so
the 95\% confidence interval of the mean is $[96.61, 97.00]\%$. The second source
is the finite size of the test set. Treating each of the $n=720$ test images as
an independent Bernoulli trial, the standard error of a single accuracy estimate
at $96.81\%$ is $0.65$ percentage points (pp), and the 95\% Wilson score interval
is $[95.25, 97.86]\%$. Combining both sources in quadrature gives a standard
uncertainty of $0.78$~pp for one support draw evaluated on this test set, so the
test-set term dominates. Because images recorded in the same acquisition session
may be correlated (section~\ref{sec:conclusion}), the effective number of
independent test samples can be smaller than 720, and these intervals should be
read as lower bounds on the true uncertainty. The expected calibration error of
the selected model is $1.04\pm0.30\%$, that is, its predicted confidences agree
closely with its observed accuracy.

Differences between methods are assessed on the same 20 support draws.
Tip-Adapter-F exceeds the training-free Tip-Adapter by $10.40$~pp (95\% confidence
interval $[9.98, 10.83]$~pp) and does so in every draw (paired $t$-test and
Wilcoxon signed-rank test, $p<0.001$). Because these paired tests condition on
the fixed test set, they capture support-draw variability only. Differences
smaller than the $0.65$~pp test-set standard error, such as the $0.57$~pp
advantage of ViT-H-14 over ViT-L-14 (section~\ref{sec:backbone}) or the $0.08$~pp
spread across the four prompt families (section~\ref{sec:prompt}), cannot be
resolved with the present test set and are therefore not interpreted as rankings.
The same applies with more force at class level, where each recall is estimated
from 120 test images: the standard errors range from $0.87$~pp (Knocking) to
$2.53$~pp (Shaking), and the 95\% Wilson interval of the Shaking recall,
$[85.2, 95.4]\%$, overlaps those of Watering and Walking. The ranking of Shaking
as the hardest class is therefore supported by its systematic confusion with
Watering (figure~\ref{fig:confusion}) and by the embedding-space adjacency of
the two classes (section~\ref{sec:umap}) rather than by its recall value alone.
All statistics in this subsection were computed from the per-draw result files.

\begin{table}[htbp]
\caption{Uncertainty budget of the selected Tip-Adapter-F model (ViT-L-14/openai,
\texttt{domain\_visual}, $K=128$) on the held-out test set. Values refer to accuracy unless stated otherwise; pp: percentage points; CI: confidence interval.}
\label{tab:uncertainty}
\centering
\begin{tabular}{llc}
\toprule
Source & Quantity & Value \\
\midrule
Support draw (20 draws) & Standard deviation & $0.42$~pp \\
 & 95\% CI of the mean & $[96.61, 97.00]\%$ \\
Finite test set ($n=720$) & Binomial standard error & $0.65$~pp \\
 & 95\% Wilson interval & $[95.25, 97.86]\%$ \\
Combined (single draw) & Standard uncertainty & $0.78$~pp \\
Per class ($n=120$) & Recall standard error & $0.87$--$2.53$~pp \\
Calibration & Expected calibration error & $1.04\pm0.30\%$ \\
\bottomrule
\end{tabular}
\end{table}

\section{Conclusion}
\label{sec:conclusion}

This study demonstrates that pre-trained vision-language representations can support
effective few-shot \Photdr{} event classification when raw multi-channel signatures
are structurally encoded into spectrogram mosaics and combined with an adapted
support cache. Direct zero-shot image--text matching is ineffective for the
present \Photdr{} representation and prompt space, since its predictions degenerate to a
chance-level accuracy, which we attribute to the substantial domain gap between
fiber-optic sensing measurements and standard natural image--text pretraining rather
than a limitation of the CLIP architecture alone.
Nevertheless, the CLIP visual features provide transferable structure that becomes
highly discriminative when exploited through Tip-Adapter-F cache adaptation.

At the practical low-label operating point of 16 shots per class, that is, only 96
labeled support images at adaptation time across the six classes, Tip-Adapter-F already reaches roughly
90\% accuracy, which we regard as the central label-efficiency finding. The objective
of this study is neither to outperform fully supervised state-of-the-art models under
abundant-label conditions, nor to match the few-shot accuracies of models that are first
pre-trained on an unlabeled in-domain \Photdr{} corpus~\cite{wang2026ssbimamba}, but to
establish how far a fixed, general-purpose vision-language representation can be taken
with no in-domain pre-training at all (Table~\ref{tab:regimes}). Scaling to
128 shots defines the peak headline performance: the validation-selected
Tip-Adapter-F model attains an accuracy of \pmv{96.81}{0.42} and a macro-F1 of
\pmv{96.79}{0.42} on a strictly isolated held-out test set. Of the three stages in Table~\ref{tab:comparison}, Tip-Adapter-F attains the highest mean accuracy, clearly surpassing both zero-shot CLIP and the training-free Tip-Adapter. Expressed as a measurement result, this accuracy carries a standard uncertainty of $0.78$~pp for a single support draw, dominated by the finite size of the test set rather than by the choice of support recordings (section~\ref{sec:uncertainty}). The prompt, backbone, and few-shot analyses delimit where the approach is robust: performance is stable across Vision Transformer encoders at high shot counts, ResNet backbones are weaker from 8 shots onwards, and at 128 shots the fused model is essentially insensitive to prompt wording, whether simple or domain-specific. Taken
together, these results present a label-efficient adaptation strategy for \Photdr{} event
classification.

One limitation concerns session independence. Because the benchmark exposes no
acquisition-session identifiers, images recorded within the same session may fall on
opposite sides of the split. Any residual within-session similarity would therefore
inflate the held-out estimates, and the reported accuracies should be read as an upper bound with respect to session-level generalization.

Two further limitations concern the evaluation design. First, the contribution of the
language branch is bounded only indirectly, by the degenerate zero-shot result and by
the insensitivity of the fused model to prompt wording, rather than isolated by a
controlled ablation of the text branch. Second, the held-out set is a fixed subsample
of 120 images per class rather than the benchmark's test partition in full. The
subsample is drawn deterministically and never touched during selection, so it
introduces no leakage, but it is smaller than the available partition; because both
encoders stay frozen, repeating the final evaluation on the complete partition requires one further embedding pass and no retraining.

Several directions remain for future work: evaluation across full train/test partitions
and cross-site or cross-session deployment, ablation of the mosaic layout and STFT
parameters, robustness to noise and amplitude drift on the raw signal streams, and
pair-aware or physics-informed strategies for the Shaking--Watering confusion; the strong
ViT-H-14 result further motivates a validation-first study of larger CLIP backbones.
The clearest bridge to the current state of the art would be to combine the two
supervision regimes of Table~\ref{tab:regimes}, that is, to continue pre-training the
CLIP image encoder on an unlabeled \Photdr{} corpus before the cache is fitted, which
would test directly whether the accuracy gap to the domain-pre-trained
models~\cite{wang2026ssbimamba} is attributable to the missing in-domain stage rather
than to the fixed-cache adaptation itself.

In summary, the results suggest that few-shot VLM adaptation can be a practical and
label-efficient option for \Photdr{} event classification when combined with careful
signal-to-image conversion and strict validation/test separation.
 
%%%%%%%%%%%%%%%%%%%%%%%%%%%%%%%%%%%%%%%%%%%%%%%%%%%%%
 
\FloatBarrier
\section*{Acknowledgments}
\textbf{Use of generative AI tools.} During the preparation of this work, the authors
used Claude (Anthropic) and ChatGPT (OpenAI) to enhance the clarity and coherence of the manuscript. The authors
reviewed, validated and critically assessed all AI-assisted outputs and take full responsibility for
the content of the manuscript.

\textbf{Competing interests.} The authors declare no competing interests.

\section*{Data availability}
The raw $\Phi$-OTDR event dataset is the publicly available benchmark released by
Cao et al.~\cite{cao2023} and is available from its authors' repository~\cite{cao2023data}. The code of this study and the corresponding
spectrogram-mosaic image dataset are shared on GitHub:
\url{https://github.com/miralab-ai/vlm-phase-otdr}. The release comprises the image-generation pipeline, the subset-selection and support-sampling seeds, the train-side/held-out split indices, the validation-selected configuration, and the full
hyperparameter-optimization and result tables.

%% ---- References ----
\FloatBarrier

\end{document}